\pdfoutput=1

\documentclass[11pt]{article}

\usepackage{acl}

\usepackage{times}
\usepackage{latexsym}

\usepackage[T1]{fontenc}

\usepackage[utf8]{inputenc}

\usepackage{microtype}

\usepackage{inconsolata}

\title{Do Zombies \emph{Understand}?\\ A Choose-Your-Own-Adventure Exploration of Machine Cognition}

\author{
    Ariel Goldstein $\qquad$
    Gabriel Stanovsky\\
    The Hebrew University of Jerusalem \\ 
    {\texttt{\{ariel.y.goldstein, gabriel.stanovsky\}@mail.huji.ac.il}}
    }

\usepackage{graphicx}
\usepackage{epigraph} 
\usepackage{tcolorbox}
\tcbuselibrary{theorems}
\newtcbtheorem[auto counter]{mytheo}{Definition}{%
lower separated=false,
                colback=white!90!gray,
colframe=white!70!black,fonttitle=\bfseries,
colbacktitle=white!70!gray,
coltitle=black,
}{th}

\usepackage[normalem]{ulem}

\definecolor{bottlegreen}{rgb}{0.0,0.42,0.31}
\definecolor{donorred}{RGB}{228.,116.,95.}
\definecolor{reciepientblue}{RGB}{0,152,251}

\newtheorem{definition}{Definition}

\newcommand{\TU}[0]{\emph{understand}}
\newcommand{\TUing}[0]{\emph{understanding}}

\newcommand{\myquote}[2]{\epigraph{#1}{#2}}

\begin{document}
\maketitle
\begin{abstract}
Recent advances in LLMs have sparked a debate on whether they \TU{} text. In this position paper, we argue that opponents in this debate hold different definitions for \TUing{}, and particularly differ in their view on the role of consciousness. To substantiate this claim, we propose a thought experiment involving an open-source chatbot $Z$ which excels on every possible benchmark, seemingly without  subjective experience. We ask whether $Z$ is capable of \TUing{}, and show that different schools of thought within seminal AI research seem to answer this question differently, uncovering their terminological disagreement. Moving forward, we propose two distinct working definitions for \TUing{} which explicitly acknowledge  the question of consciousness, and draw connections with a rich literature in philosophy, psychology and neuroscience.

\end{abstract}

\section{Introduction: A Thought Experiment}



Large language models (LLMs) achieve impressive results on various benchmarks, seeming to generalize to unseen tasks and domains~\citep{brown2020language}. This initiated a debate on whether LLMs truly \TU{}~\citep{Mitchell2022TheDO}. On the one hand, several works claim that LLMs are starting to show signs of understanding text~\cite{Manning2022HumanLU,Piantadosi2022MeaningWR,bubeck2023sparks}, while on the other hand, others 
argue that LLMs are inherently incapable of understanding because they observe form without meaning~\cite{bender-koller-2020-climbing,bender2021dangers,marcusnonsense}. 
Evidently, such works have differing opinions of what it means for a model to \TU{}.
Here, we do not advocate for a single ``true'' definition for \TUing{}, and instead aim 
to shed new light on the roots of this debate.

We contextualize the debate on machine cognition within the \emph{mind-body problem},
which has been at the center of vast philosophical debate, as well as intense empirical research in cognitive neuroscience. 
We follow \citet{chalmers1995facing}, who asks whether the quality of \emph{consciousness} - the ability to have subjective experiences - is a strict requirement for \TUing{}, 
or whether 
it can also manifest in non-conscious agents. 
We argue that this question lies in the background of all discussion around whether LLMs truly \TU{}.

To make this concrete, consider the following thought experiment: you are presented with $Z$, a newfangled chatbot.
$Z$ is implemented in computer hardware and performs only mathematical manipulations of its input. It is completely open-source --- you have access to its code, training data, weights, hyperparmeters, and any other implementation detail. You interact with $Z$ and discover that it excels on all NLP benchmarks, and will do so on any possible test you will come up with in the future. 
In essence, $Z$ is the chatbot equivalent of the philosophical zombie~\citep{Kirk1974SentienceAB,Chalmers1996TheCM}; it outperforms humans on all tasks, supposedly without having subjective experience.
\emph{Do you consider $Z$ as capable of} ``\TUing{}''?

If you answer \emph{``Yes''}, turn to Section~\ref{sec:zombies-understand}. For you, the path toward machine cognition lies in test sets of ever-increasing complexity, identifying ever-more subtle deficiencies in machine responses. If we reach this road's end, we will find $Z$.

If you answer \emph{``No''}, turn to Section~\ref{sec:zombies-dont-understand}. You hold that consciousness is a prerequisite for \TUing{},  as that is the only thing distinguishing zombies from humans.
We make several connections between recent neuroscience research and AI, e.g., the function of consciousness and advancements in the field of neural correlates of consciousness.

If you feel 
uncomfortable with either of these options,
 turn to Section~\ref{sec:other-answers},
where we address potential objections to our setup and assumptions.

This setup produces two distinct definitions and research agendas for machine \TUing{}, which are currently conflated in AI discussion.

\section{Background: Philosophical Zombies}
\label{sec:background}
\label{sec:bg}
The zombie argument is a thought experiment proposed in the context of debates about consciousness and its relationship to the physical world, i.e., to what is measurable~\citep{Kirk1974SentienceAB}. It seeks to question the validity of physicalism, the belief that all that exists in our world, including consciousness, is physical~\citep{sep-physicalism}. The zombie argument suggests that it is conceptually possible for there to be beings that are physically identical to humans but possess no conscious experiences. These are commonly referred to as ``philosophical zombies''.

Philosophical zombies behave just like humans. They appear to feel pain when injured or joy when pleased, and can converse about the events in which they participate. Despite these behaviors, philosophical zombies possess no subjective experiences or ``qualia''~\citep{sep-qualia} – they do not consciously experience sensations, feelings, or thoughts. For instance, a philosophical zombie would react externally like a human would to stepping on a sharp object but would not internally suffer due to the painful sensation. \citet{Chalmers1996TheCM} played a significant role in bringing the argument into the mainstream discourse, particularly in the context of the philosophy of mind. 

We conjure the equivalent of a zombie chatbot. It is implemented on physical computer hardware, and it is capable of excelling on every NLP task, seemingly without conscious experience.

\section{Zombies \textit{do} Understand:\\ Functional Definition of Understanding}
\label{sec:zombies-understand}

One approach to machine cognition relies only on the model's behavior, independent of any internal experience. This definition holds that \TUing{} can be inferred from performance on specific tasks. We formulate this notion for a task $T$ in Definition~\ref{th:func-understanding}: 

\begin{definition}{Functional Understanding.}
\label{th:func-understanding}
     A model $Z$ functionally understands a task $T$ if its performance on $T$ is as good (or better than) a human  who is an expert in $T$.
\end{definition}



This approach to \TU{}ing is articulated in \citet{dummett1996theory}'s discussion around intelligence:\footnote{Emphasis is our own in all quotes.}

\myquote{If a Martian could learn to speak a human language, or \textbf{a robot be devised to behave in just the ways that are essential to a language speaker}, an implicit knowledge of the correct theory of meaning for the language could be attributed to the Martian or the robot with as much right as to a human speaker, even though their internal mechanisms were entirely different.}{\citet{dummett1996theory}}

This framing helps explain the common practice for testing \TU{}ing in models through longstanding challenges, such as chess, Go, or language generation, or in many NLP benchmarks, such as text comprehension~\citep{wang-etal-2018-glue,hendrycks2021measuring,srivastava2023beyond,liang2023holistic} or formal semantic representation~\citep{oepen2014semeval,Nivre2016UniversalDV}. 

\citet{McCarthy1990ChessAT} figuratively called such tasks the \emph{Drosophila of AI}, drawing a parallel between research in AI and biology, where model organisms (e.g., the Drosophila fly) are chosen for wide benchmark experimentation with findings generalizing beyond that specific organism.

Evidently, the recurring trend in the last 70 years has seen tasks adopted as benchmarks for \TUing{} until automated models functionally understand them. Then, the AI community \emph{moves the goalposts} to another, arguably harder, external objective benchmark for \TUing{}. 
Taken to the extreme, a model that functionally understands \emph{every} potential benchmark is equivalent to our hypothetical $Z$ chatbot. 
Notably, models excelling on these tasks are tested only externally and are not required to have any internal state linked to their success.
Below, we outline some famous examples of this trend.



Perhaps the most well-known examples are the games of chess and Go. Chess served as a proxy task for \TUing{} for nearly 50 years. Early works, such as
\citet{shannon1950programming} and \citet{turing1953digital}, already deemed chess 
a benchmark for machine intelligence.
With the advent of deep learning models, chess engines now vastly outplay any human opponent~\citep{Silver2017MasteringCA}.
For all intents and purposes, these models  \emph{functionally understand} chess according to Definition~\ref{th:func-understanding}.
Consequently, chess was abandoned as a useful benchmark for \TUing{}.\footnote{New chess engines are still being developed, albeit without any claims about general \TUing{} beyond chess.}
Instead, Go was adopted as a marker for \TUing{}~\cite{bouzy2001computer,van2004ai}, until Go models outplayed the best human players~\cite{silver2016mastering}. 

\begin{figure}[tb!]
        \centering
        \includegraphics[width=0.47\textwidth]{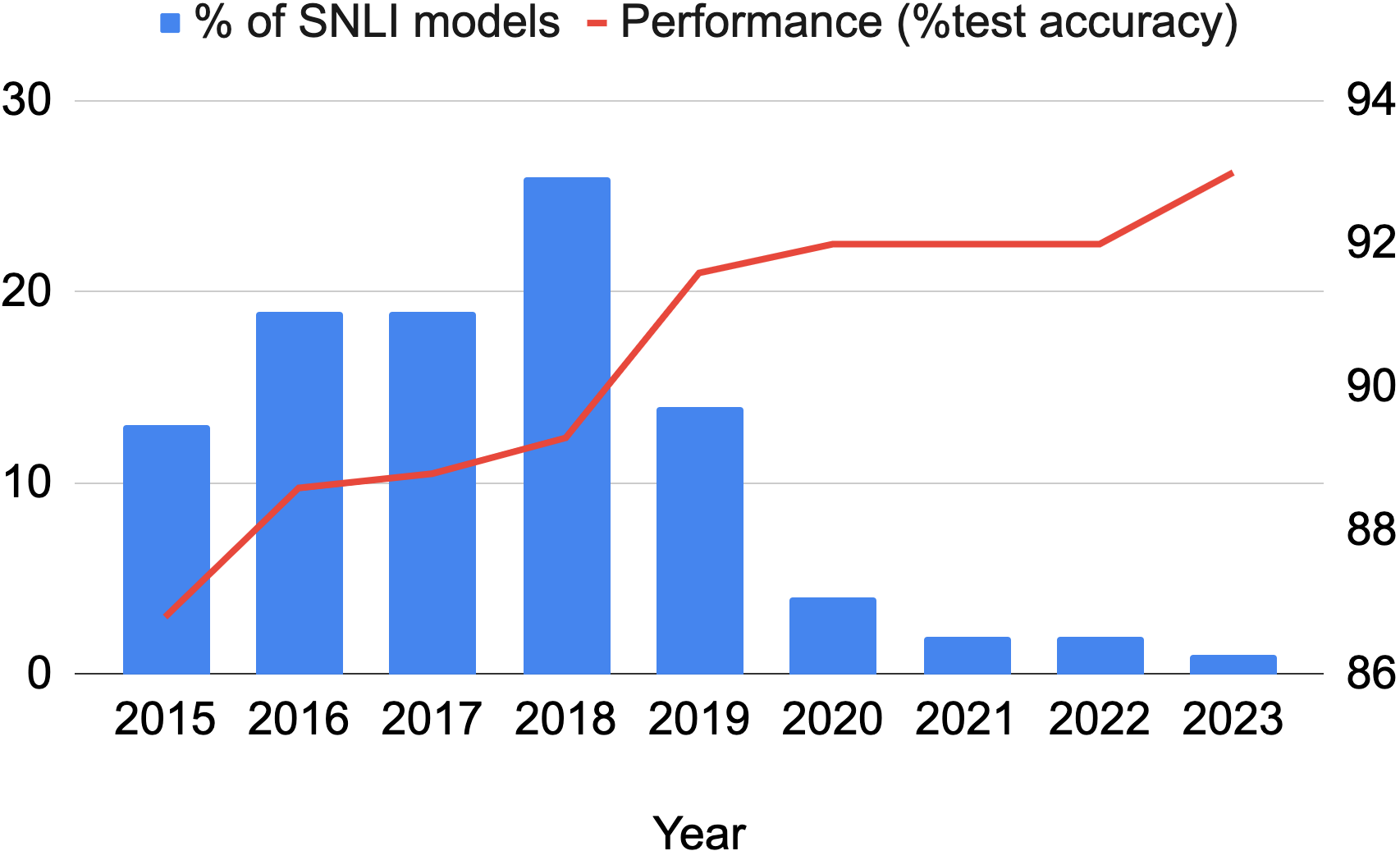}
        \caption{\label{fig:snli} \%Models tested on SNLI (blue bars, left axis) per year versus state-of-the-art performance on the benchmark (red line, right axis). Data collected from \url{paperswithcode.com}.}
\end{figure}

A natural follow-up question is whether LLMs  can functionally understand. 
We argue that similar trends to Go and chess happen for certain NLP tasks. For example, natural language inference (NLI) has garnered significant attention since its introduction~\citep{Dagan2005ThePR}, and was framed as ``fundamental to understanding natural language'' by the authors of SNLI, one of the most prominent benchmarks for the task~\citep{Bowman2015ALA}. 
However, as can be seen in Figure~\ref{fig:snli}, the number of models developed over SNLI has dropped in recent years when performance on the benchmark was saturated, while similar trends are observed also for the follow-up MNLI dataset~\citep{N18-1101}. To the best of our knowledge, there is no large scale effort to curate a new benchmark for the task. It could be argued that LLMs functionally understand NLI, and the field has implicitly moved to other tasks. An indication that this trend does not stem from loss of interest in the task are various recent works that use NLI models as components within larger systems, showing that indeed NLI models are useful~\citep{honovich-etal-2021-q2,laban-etal-2022-summac,aharoni-etal-2023-multilingual,min2023factscore}.

Adopting this notion of \TUing{} implies getting other NLP tasks to go down this path, incrementally achieving functional understanding on as many tasks as possible. At the end of this path, \emph{if it is reachable}, lies our hypothetical $Z$ chatbot, which functionally understands \emph{every} NLP task.




\section{Zombies \emph{don't} Understand: Consciousness is a Prerequisite for Understanding}
\label{sec:zombies-dont-understand}

In contrast to the external approach to \TUing{} in AI, stands a long line of work that either explicitly or implicitly requires models to have subjective experience.
These works view the quality of consciousness as an essential aspect of \TUing{}, in addition to accurate performance on any particular task.

This notion is formulated with regards to a model $M$ and a task $T$ in the following definition:

\begin{definition}{Conscious Understanding.}
    \label{th:conc-understanding}
    $M$ \emph{consciously understands} $T$ if both hold:
    \begin{enumerate}
        \item $M$ functionally understands $T$ (\S Def.~\ref{th:func-understanding}).
        \item $M$ is conscious -- it has immediate subjective experience. In \citet{nagel1974like}'s words there is something that ``it is like'' to be $M$.
    \end{enumerate}
\end{definition}

As we highlight below, this notion of \TUing{} has been articulated by seminal works in the field of AI and NLP.
In a section titled \emph{Argument from Consciousness} from his famous paper, \citet{Turing1950ComputingMA} cites \citep{jefferson1949mind}:

\myquote{Not until a machine can write a sonnet or compose a concerto because of \textbf{thoughts and emotions felt}, and not by the chance fall of symbols, could we agree that machine equals brain—that is, not only write it but \textbf{know that it had written it}. No mechanism could feel (and not merely artificially signal, an easy contrivance) pleasure at its successes, grief when its valves fuse, be warmed by flattery, be made miserable by its mistakes, be charmed by sex, be angry or depressed when it cannot get what it wants.}{\citet{Turing1950ComputingMA}}

This reveals a strong tie between external behaviors, such as writing a sonnet or composing music, and subjective experiences, such as feeling emotions, in considering them as prerequisites for \TUing{}, or intelligence. 

Similar connection between consciousness and understanding is also evident in Searle's interpretation for his Chinese room argument~\citep{searle1980minds}. 
This questions if a computer can be truly intelligent by imagining a non-Chinese speaker using a rule-book to manipulate Chinese symbols, seemingly displaying comprehension without real \TUing{}.  \citet{searle2010dualism} explicitly states that this argument was meant as a thought experiment for the existence of consciousness, or its lack thereof:

\myquote{I demonstrated years ago with the so-called Chinese Room Argument that the implementation of the computer program is not by itself sufficient for \textbf{consciousness} or intentionality.}{\citet{searle2010dualism}}


Other notable works have also connected the Turing test and  the Chinese room argument to consciousness \cite{churchland1990could,gozzano1995consciousness,dehaene2012single}. 

We believe that consciousness also underlies the current discussion on whether LLMs \TU{}, as evident in \citep{bender2021dangers}:


\myquote{Our human understanding of coherence derives from our ability to recognize interlocutors’ \textbf{beliefs and
intentions} within context. That is, human language use takes place between individuals who share common ground and are
mutually \textbf{aware} of that sharing (and its extent), who have communicative intents which they use language to convey, and who \textbf{model
each others’ mental states as they communicate.} }{\citet{bender2021dangers}}

Finally, this definition for understanding is in line with \citep{o2021god}, who in her recent book advocated for consciousness as the defining factor of human intelligence:

\myquote{As AI continues to blow past us in benchmark after benchmark of higher cognition \textbf{we quell our anxiety by insisting that what distinguishes true consciousness is emotions, perception, the ability to experience and feel}. The qualities, in other words, which we share with animals.}{\citet{o2021god}}


To move forward on conscious understanding as articulated in Definition~\ref{th:conc-understanding}, we suggest following literature in psychology and neuroscience regarding tests for consciousness (for review, see \cite{bayne2024tests}) and specifically the neural-correlate-of-consciousness~(NCC;~\citealp{Koch2016NeuralCO}). This field is dedicated to recognizing the neural dynamics in biological organisms associated with consciousness experience. For example, the Integrated Information Theory~(IIT;~\citealp{Tononi2016IntegratedIT}), specifically the weak IIT, links elements of consciousness with wider information flow metrics, like recurrent processing or global workspace. 
These findings can inform cognitively-inspired architectures, e.g., spiking neural networks~\citep{Mediano2022TheSO}.


\section{Other Possible Answers}
\label{sec:other-answers}
Here we survey alternative answers to the question of whether zombies \TU{}. 
We reply to these objections below, hopefully resolving seeming inconsistencies within our paradigm. 

\paragraph{Argument:} \emph{Whether $Z$} \TU{}\emph{s \textbf{depends on its implementation} (training data, architecture, hyperparameters, etc.), but it has nothing to do with conscious experience.}

This argument is in line with \citet{block1981psychologism}'s definition of \emph{Psychologism}, which assumes that there may exist implementations of $Z$ which will show that it indeed \TU{}\emph{s}, e.g., if they involve complex feature manipulations or explicit  reasoning steps, while there may exist other implementations which imply that $Z$ does not \TU{}, e.g., if all $Z$ does is leverage spurious correlations or memorize an immense look up table, similar to the Chinese room argument~\citep{searle1980minds}. 

We argue that the concerns regarding specific implementations not being indicative of \TUing{} can be mitigated with our requirement that $Z$ excels on all possible NLP benchmarks, while also being implemented on a physical hardware.
For example, if $Z$ leverages spurious correlations, then by definition there are samples which do not exhibit these correlations and which will stump $Z$ (otherwise they would not be spurious), contradicting our assumption that $Z$ is a philosophical zombie, and does not make non-human errors. 
Similarly, since human language can produce an infinite amount of meaningful texts~\citep{chomsky2002syntactic}, and $Z$ can only memorize a finite amount of samples (as it is implemented in finite hardware), then there must be samples outside of its memory on which it is bound to fail. This again contradicts our assumption that $Z$ does not fail where humans do not fail. 



\paragraph{Argument:}\emph{The question is ill-posed as \textbf{$Z$ is inconceivable}. Hence it is meaningless to discuss different properties of $Z$.}

This argument may stem from the belief that consciousness has a function in \TUing{}~\citep{sep-consciousness}, and hence it is impossible for an agent to excel on every NLP benchmark without also achieving consciousness. 
We argue that this position is  compatible with the view that consciousness is a prerequisite for \TUing{}~(\S\ref{sec:zombies-dont-understand}), by positing that is in fact needed to achieve functional understanding.

\paragraph{Argument:} \emph{The question is ill-posed as \textbf{it does not define what is}} \textbf{\TUing{}}\emph{. Different definitions may lead to different answers.}

We do not aim to define apriori what constitutes \TUing{}, and do not argue that there is a single ``correct'' definition. 
Instead, we try to tease apart what researchers mean when they use the term, 
specifically highlighting the role that consciousness plays in it, and examine how AI research may be explained through this lens. In fact, we claim that answering the question elucidates different definitions for \TUing{}~(Definitions~\ref{th:func-understanding},\ref{th:conc-understanding}). We invite researchers to engage with this question to examine \emph{their} definition for \TUing{}.


\section{Discussion}
We pose the question of whether Zombies \TU{} to highlight consciousness's role in AI debates. We propose two definitions for \TUing{}. One deals with \emph{functional understanding}, and the other revolves around \emph{conscious experience}. These definitions give rise to different research agendas.
This argument can be ported to other discussions about  LLMs possessing human traits. E.g., \citet{Perry2023AIWN} recently claimed that LLMs could not feel empathy. We argue that here, too, consciousness plays a major role in the definition of empathy. 
Similarly, the question of the relevance of consciousness to empathy can be unpacked by asking ``Can Zombies be \emph{Empathetic}?''.

\section*{Limitations}
We presented a thought experiment posing a philosophical question and have tried to answer it through the lens of two schools of thought within the fields of AI and NLP. While we tried to address potential reservations to our paradigm, it is possible that there are other answers that were not considered in this paper. We invite  opinions and objections to further inform the discussion around machine cognition.

\section*{Acknowledgments}
We would like to thank the anonymous reviewers for their thoughtful comments and suggestions, and Dr. Anat Arzi, Dr. Yael Bitterman, and 
Prof. Oron Shagrir, for many insightful and productive discussions. This work was partially supported by a grant from the Israeli Ministry of Science and Technology (grant no. 2336).









\bibliography{anthology,custom}

\begin{thebibliography}{51}
\expandafter\ifx\csname natexlab\endcsname\relax\def\natexlab#1{#1}\fi

\bibitem[{Aharoni et~al.(2023)Aharoni, Narayan, Maynez, Herzig, Clark, and
  Lapata}]{aharoni-etal-2023-multilingual}
Roee Aharoni, Shashi Narayan, Joshua Maynez, Jonathan Herzig, Elizabeth Clark,
  and Mirella Lapata. 2023.
\newblock \href {https://doi.org/10.18653/v1/2023.findings-acl.220}
  {Multilingual summarization with factual consistency evaluation}.
\newblock In \emph{Findings of the Association for Computational Linguistics:
  ACL 2023}, pages 3562--3591, Toronto, Canada. Association for Computational
  Linguistics.

\bibitem[{Bayne et~al.(2024)Bayne, Seth, Massimini, Shepherd, Cleeremans,
  Fleming, Malach, Mattingley, Menon, Owen et~al.}]{bayne2024tests}
Tim Bayne, Anil~K Seth, Marcello Massimini, Joshua Shepherd, Axel Cleeremans,
  Stephen~M Fleming, Rafael Malach, Jason~B Mattingley, David~K Menon, Adrian~M
  Owen, et~al. 2024.
\newblock Tests for consciousness in humans and beyond.
\newblock \emph{Trends in Cognitive Sciences}.

\bibitem[{bench authors(2023)}]{srivastava2023beyond}
BIG bench authors. 2023.
\newblock \href {https://openreview.net/forum?id=uyTL5Bvosj} {Beyond the
  imitation game: Quantifying and extrapolating the capabilities of language
  models}.
\newblock \emph{Transactions on Machine Learning Research}.

\bibitem[{Bender et~al.(2021)Bender, Gebru, McMillan-Major, and
  Shmitchell}]{bender2021dangers}
Emily~M Bender, Timnit Gebru, Angelina McMillan-Major, and Shmargaret
  Shmitchell. 2021.
\newblock On the dangers of stochastic parrots: Can language models be too big?
\newblock In \emph{Proceedings of the 2021 ACM conference on fairness,
  accountability, and transparency}, pages 610--623.

\bibitem[{Bender and Koller(2020)}]{bender-koller-2020-climbing}
Emily~M. Bender and Alexander Koller. 2020.
\newblock \href {https://doi.org/10.18653/v1/2020.acl-main.463} {Climbing
  towards {NLU}: {On} meaning, form, and understanding in the age of data}.
\newblock In \emph{Proceedings of the 58th Annual Meeting of the Association
  for Computational Linguistics}, pages 5185--5198, Online. Association for
  Computational Linguistics.

\bibitem[{Block(1981)}]{block1981psychologism}
Ned Block. 1981.
\newblock Psychologism and behaviorism.
\newblock \emph{Philosophical Review}, 90(1):5--43.

\bibitem[{Bouzy and Cazenave(2001)}]{bouzy2001computer}
Bruno Bouzy and Tristan Cazenave. 2001.
\newblock Computer go: an ai oriented survey.
\newblock \emph{Artificial Intelligence}, 132(1):39--103.

\bibitem[{Bowman et~al.(2015)Bowman, Angeli, Potts, and
  Manning}]{Bowman2015ALA}
Samuel~R. Bowman, Gabor Angeli, Christopher Potts, and Christopher~D. Manning.
  2015.
\newblock \href {https://doi.org/10.18653/v1/D15-1075} {A large annotated
  corpus for learning natural language inference}.
\newblock In \emph{Proceedings of the 2015 Conference on Empirical Methods in
  Natural Language Processing}, pages 632--642, Lisbon, Portugal. Association
  for Computational Linguistics.

\bibitem[{Brown et~al.(2020)Brown, Mann, Ryder, Subbiah, Kaplan, Dhariwal,
  Neelakantan, Shyam, Sastry, Askell, Agarwal, Herbert{-}Voss, Krueger,
  Henighan, Child, Ramesh, Ziegler, Wu, Winter, Hesse, Chen, Sigler, Litwin,
  Gray, Chess, Clark, Berner, McCandlish, Radford, Sutskever, and
  Amodei}]{brown2020language}
Tom~B. Brown, Benjamin Mann, Nick Ryder, Melanie Subbiah, Jared Kaplan,
  Prafulla Dhariwal, Arvind Neelakantan, Pranav Shyam, Girish Sastry, Amanda
  Askell, Sandhini Agarwal, Ariel Herbert{-}Voss, Gretchen Krueger, Tom
  Henighan, Rewon Child, Aditya Ramesh, Daniel~M. Ziegler, Jeffrey Wu, Clemens
  Winter, Christopher Hesse, Mark Chen, Eric Sigler, Mateusz Litwin, Scott
  Gray, Benjamin Chess, Jack Clark, Christopher Berner, Sam McCandlish, Alec
  Radford, Ilya Sutskever, and Dario Amodei. 2020.
\newblock \href
  {https://proceedings.neurips.cc/paper/2020/hash/1457c0d6bfcb4967418bfb8ac142f64a-Abstract.html}
  {Language models are few-shot learners}.
\newblock In \emph{Advances in Neural Information Processing Systems 33: Annual
  Conference on Neural Information Processing Systems 2020, NeurIPS 2020,
  December 6-12, 2020, virtual}.

\bibitem[{Bubeck et~al.(2023)Bubeck, Chandrasekaran, Eldan, Gehrke, Horvitz,
  Kamar, Lee, Lee, Li, Lundberg, Nori, Palangi, Ribeiro, and
  Zhang}]{bubeck2023sparks}
Sébastien Bubeck, Varun Chandrasekaran, Ronen Eldan, Johannes Gehrke, Eric
  Horvitz, Ece Kamar, Peter Lee, Yin~Tat Lee, Yuanzhi Li, Scott Lundberg,
  Harsha Nori, Hamid Palangi, Marco~Tulio Ribeiro, and Yi~Zhang. 2023.
\newblock \href {http://arxiv.org/abs/2303.12712} {Sparks of artificial general
  intelligence: Early experiments with gpt-4}.

\bibitem[{Chalmers(1996)}]{Chalmers1996TheCM}
D.~Chalmers. 1996.
\newblock \href {https://api.semanticscholar.org/CorpusID:142058675} {The
  conscious mind: in search of a fundamental theory}.

\bibitem[{Chalmers(1995)}]{chalmers1995facing}
David~J Chalmers. 1995.
\newblock Facing up to the problem of consciousness.
\newblock \emph{Journal of consciousness studies}, 2(3):200--219.

\bibitem[{Chomsky(2002)}]{chomsky2002syntactic}
Noam Chomsky. 2002.
\newblock \emph{Syntactic structures}.
\newblock Mouton de Gruyter.

\bibitem[{Churchland and Churchland(1990)}]{churchland1990could}
Paul~M Churchland and Patricia~Smith Churchland. 1990.
\newblock Could a machine think?
\newblock \emph{Scientific American}, 262(1):32--39.

\bibitem[{Dagan et~al.(2005)Dagan, Glickman, and Magnini}]{Dagan2005ThePR}
Ido Dagan, Oren Glickman, and Bernardo Magnini. 2005.
\newblock \href {https://api.semanticscholar.org/CorpusID:8587959} {The pascal
  recognising textual entailment challenge}.
\newblock In \emph{Machine Learning Challenges Workshop}.

\bibitem[{Dehaene and Sigman(2012)}]{dehaene2012single}
Stanislas Dehaene and Mariano Sigman. 2012.
\newblock From a single decision to a multi-step algorithm.
\newblock \emph{Current opinion in neurobiology}, 22(6):937--945.

\bibitem[{Dummett(1996)}]{dummett1996theory}
Michael Dummett. 1996.
\newblock What is a theory of meaning?(i).
\newblock \emph{The seas of language}, pages 1--33.

\bibitem[{Gozzano(1995)}]{gozzano1995consciousness}
Simone Gozzano. 1995.
\newblock Consciousness and understanding in the chinese room.

\bibitem[{Hendrycks et~al.(2021)Hendrycks, Burns, Basart, Zou, Mazeika, Song,
  and Steinhardt}]{hendrycks2021measuring}
Dan Hendrycks, Collin Burns, Steven Basart, Andy Zou, Mantas Mazeika, Dawn
  Song, and Jacob Steinhardt. 2021.
\newblock \href {https://openreview.net/forum?id=d7KBjmI3GmQ} {Measuring
  massive multitask language understanding}.
\newblock In \emph{9th International Conference on Learning Representations,
  {ICLR} 2021, Virtual Event, Austria, May 3-7, 2021}. OpenReview.net.

\bibitem[{Honovich et~al.(2021)Honovich, Choshen, Aharoni, Neeman, Szpektor,
  and Abend}]{honovich-etal-2021-q2}
Or~Honovich, Leshem Choshen, Roee Aharoni, Ella Neeman, Idan Szpektor, and Omri
  Abend. 2021.
\newblock \href {https://doi.org/10.18653/v1/2021.emnlp-main.619} {$q^{2}$:
  {E}valuating factual consistency in knowledge-grounded dialogues via question
  generation and question answering}.
\newblock In \emph{Proceedings of the 2021 Conference on Empirical Methods in
  Natural Language Processing}, pages 7856--7870, Online and Punta Cana,
  Dominican Republic. Association for Computational Linguistics.

\bibitem[{Jefferson(1949)}]{jefferson1949mind}
Geoffrey Jefferson. 1949.
\newblock The mind of mechanical man.
\newblock \emph{British Medical Journal}, 1(4616):1105.

\bibitem[{Kirk(1974)}]{Kirk1974SentienceAB}
Robert Kirk. 1974.
\newblock \href {https://api.semanticscholar.org/CorpusID:170114519} {Sentience
  and behaviour}.
\newblock \emph{Mind}, pages 43--60.

\bibitem[{Koch et~al.(2016)Koch, Massimini, Boly, and
  Tononi}]{Koch2016NeuralCO}
Christof Koch, Marcello Massimini, M{\'e}lanie Boly, and Giulio Tononi. 2016.
\newblock \href {https://api.semanticscholar.org/CorpusID:5395332} {Neural
  correlates of consciousness: progress and problems}.
\newblock \emph{Nature Reviews Neuroscience}, 17:307--321.

\bibitem[{Laban et~al.(2022)Laban, Schnabel, Bennett, and
  Hearst}]{laban-etal-2022-summac}
Philippe Laban, Tobias Schnabel, Paul~N. Bennett, and Marti~A. Hearst. 2022.
\newblock \href {https://doi.org/10.1162/tacl_a_00453} {{S}umma{C}: Re-visiting
  {NLI}-based models for inconsistency detection in summarization}.
\newblock \emph{Transactions of the Association for Computational Linguistics},
  10:163--177.

\bibitem[{Liang et~al.(2023)Liang, Bommasani, Lee, Tsipras, Soylu, Yasunaga,
  Zhang, Narayanan, Wu, Kumar et~al.}]{liang2023holistic}
Percy Liang, Rishi Bommasani, Tony Lee, Dimitris Tsipras, Dilara Soylu,
  Michihiro Yasunaga, Yian Zhang, Deepak Narayanan, Yuhuai Wu, Ananya Kumar,
  et~al. 2023.
\newblock Holistic evaluation of language models.
\newblock \emph{Transactions on Machine Learning Research}.

\bibitem[{Manning(2022)}]{Manning2022HumanLU}
Christopher~D. Manning. 2022.
\newblock \href {https://api.semanticscholar.org/CorpusID:248377870} {Human
  language understanding \& reasoning}.
\newblock \emph{Daedalus}, 151:127--138.

\bibitem[{Marcus(2022)}]{marcusnonsense}
Gary Marcus. 2022.
\newblock \href {https://garymarcus.substack.com/p/nonsense-on-stilts}
  {Nonsense on stilts}.

\bibitem[{McCarthy(1990)}]{McCarthy1990ChessAT}
John McCarthy. 1990.
\newblock Chess as the drosophila of ai.

\bibitem[{Mediano et~al.(2022)Mediano, Rosas, Bor, Seth, and
  Barrett}]{Mediano2022TheSO}
Pedro A.~M. Mediano, Fernando~E. Rosas, Daniel Bor, Anil.~K. Seth, and Adam~B.
  Barrett. 2022.
\newblock \href {https://api.semanticscholar.org/CorpusID:249245931} {The
  strength of weak integrated information theory}.
\newblock \emph{Trends in Cognitive Sciences}, 26:646--655.

\bibitem[{Min et~al.(2023)Min, Krishna, Lyu, Lewis, Yih, Koh, Iyyer,
  Zettlemoyer, and Hajishirzi}]{min2023factscore}
Sewon Min, Kalpesh Krishna, Xinxi Lyu, Mike Lewis, Wen-tau Yih, Pang~Wei Koh,
  Mohit Iyyer, Luke Zettlemoyer, and Hannaneh Hajishirzi. 2023.
\newblock \href {https://arxiv.org/abs/2305.14251} {Factscore: Fine-grained
  atomic evaluation of factual precision in long form text generation}.
\newblock \emph{ArXiv preprint}, abs/2305.14251.

\bibitem[{Mitchell and Krakauer(2022)}]{Mitchell2022TheDO}
Melanie Mitchell and David~C. Krakauer. 2022.
\newblock \href {https://api.semanticscholar.org/CorpusID:253107905} {The
  debate over understanding in ai’s large language models}.
\newblock \emph{Proceedings of the National Academy of Sciences of the United
  States of America}, 120.

\bibitem[{Nagel(1974)}]{nagel1974like}
Thomas Nagel. 1974.
\newblock What is it like to be a bat?
\newblock \emph{The philosophical review}, 83(4):435--450.

\bibitem[{Nivre et~al.(2016)Nivre, de~Marneffe, Ginter, Goldberg, Haji{\v{c}},
  Manning, McDonald, Petrov, Pyysalo, Silveira, Tsarfaty, and
  Zeman}]{Nivre2016UniversalDV}
Joakim Nivre, Marie-Catherine de~Marneffe, Filip Ginter, Yoav Goldberg, Jan
  Haji{\v{c}}, Christopher~D. Manning, Ryan McDonald, Slav Petrov, Sampo
  Pyysalo, Natalia Silveira, Reut Tsarfaty, and Daniel Zeman. 2016.
\newblock \href {https://aclanthology.org/L16-1262} {{U}niversal {D}ependencies
  v1: A multilingual treebank collection}.
\newblock In \emph{Proceedings of the Tenth International Conference on
  Language Resources and Evaluation ({LREC}'16)}, pages 1659--1666,
  Portoro{\v{z}}, Slovenia. European Language Resources Association (ELRA).

\bibitem[{Oepen et~al.(2014)Oepen, Kuhlmann, Miyao, Zeman, Flickinger,
  Haji{\v{c}}, Ivanova, and Zhang}]{oepen2014semeval}
Stephan Oepen, Marco Kuhlmann, Yusuke Miyao, Daniel Zeman, Dan Flickinger, Jan
  Haji{\v{c}}, Angelina Ivanova, and Yi~Zhang. 2014.
\newblock \href {https://doi.org/10.3115/v1/S14-2008} {{S}em{E}val 2014 task 8:
  Broad-coverage semantic dependency parsing}.
\newblock In \emph{Proceedings of the 8th International Workshop on Semantic
  Evaluation ({S}em{E}val 2014)}, pages 63--72, Dublin, Ireland. Association
  for Computational Linguistics.

\bibitem[{O'Gieblyn(2021)}]{o2021god}
Meghan O'Gieblyn. 2021.
\newblock \emph{God, human, animal, machine: Technology, metaphor, and the
  search for meaning}.
\newblock Anchor.

\bibitem[{Perry(2023)}]{Perry2023AIWN}
Anat Perry. 2023.
\newblock \href {https://api.semanticscholar.org/CorpusID:259995103} {Ai will
  never convey the essence of human empathy}.
\newblock \emph{Nature Human Behaviour}, 7:1808 -- 1809.

\bibitem[{Piantadosi and Hill(2022)}]{Piantadosi2022MeaningWR}
Steven~T. Piantadosi and Felix Hill. 2022.
\newblock \href {https://arxiv.org/abs/2208.02957} {Meaning without reference
  in large language models}.
\newblock \emph{ArXiv preprint}, abs/2208.02957.

\bibitem[{Searle(2010)}]{searle2010dualism}
John Searle. 2010.
\newblock Why dualism (and materialism) fail to account for consciousness.
\newblock \emph{Questioning nineteenth century assumptions about knowledge,
  III: Dualism}, pages 5--48.

\bibitem[{Searle(1980)}]{searle1980minds}
John~R Searle. 1980.
\newblock Minds, brains, and programs.
\newblock \emph{Behavioral and brain sciences}, 3(3):417--424.

\bibitem[{Shannon(1950)}]{shannon1950programming}
Claude~E Shannon. 1950.
\newblock Programming a computer for playing chess.
\newblock \emph{Philosophical Magazine}, 41(314):256--275.

\bibitem[{Silver et~al.(2016)Silver, Huang, Maddison, Guez, Sifre, Van
  Den~Driessche, Schrittwieser, Antonoglou, Panneershelvam, Lanctot
  et~al.}]{silver2016mastering}
David Silver, Aja Huang, Chris~J Maddison, Arthur Guez, Laurent Sifre, George
  Van Den~Driessche, Julian Schrittwieser, Ioannis Antonoglou, Veda
  Panneershelvam, Marc Lanctot, et~al. 2016.
\newblock Mastering the game of go with deep neural networks and tree search.
\newblock \emph{nature}, 529(7587):484--489.

\bibitem[{Silver et~al.(2017)Silver, Hubert, Schrittwieser, Antonoglou, Lai,
  Guez, Lanctot, Sifre, Kumaran, Graepel, Lillicrap, Simonyan, and
  Hassabis}]{Silver2017MasteringCA}
David Silver, Thomas Hubert, Julian Schrittwieser, Ioannis Antonoglou, Matthew
  Lai, Arthur Guez, Marc Lanctot, L.~Sifre, Dharshan Kumaran, Thore Graepel,
  Timothy~P. Lillicrap, Karen Simonyan, and Demis Hassabis. 2017.
\newblock \href {https://arxiv.org/abs/1712.01815} {Mastering chess and shogi
  by self-play with a general reinforcement learning algorithm}.
\newblock \emph{ArXiv preprint}, abs/1712.01815.

\bibitem[{Stoljar(2024)}]{sep-physicalism}
Daniel Stoljar. 2024.
\newblock {Physicalism}.
\newblock In Edward~N. Zalta and Uri Nodelman, editors, \emph{The {Stanford}
  Encyclopedia of Philosophy}, {S}pring 2024 edition. Metaphysics Research Lab,
  Stanford University.

\bibitem[{Tononi et~al.(2016)Tononi, Boly, Massimini, and
  Koch}]{Tononi2016IntegratedIT}
Giulio Tononi, M{\'e}lanie Boly, Marcello Massimini, and Christof Koch. 2016.
\newblock \href {https://api.semanticscholar.org/CorpusID:21347087} {Integrated
  information theory: from consciousness to its physical substrate}.
\newblock \emph{Nature Reviews Neuroscience}, 17:450--461.

\bibitem[{Turing(1953)}]{turing1953digital}
Alan Turing. 1953.
\newblock Digital computers applied to games.
\newblock In B.~V. Bowden, editor, \emph{Faster than thought}, pages 286--310.
  Sir Isaac Pitman \& Sons, Ltd., London.

\bibitem[{Turing(1950)}]{Turing1950ComputingMA}
Alan~M. Turing. 1950.
\newblock Computing machinery and intelligence.
\newblock \emph{Mind}, LIX:433--460.

\bibitem[{Tye(2021)}]{sep-qualia}
Michael Tye. 2021.
\newblock {Qualia}.
\newblock In Edward~N. Zalta, editor, \emph{The {Stanford} Encyclopedia of
  Philosophy}, {F}all 2021 edition. Metaphysics Research Lab, Stanford
  University.

\bibitem[{Van Der~Werf(2004)}]{van2004ai}
Erik Van Der~Werf. 2004.
\newblock \emph{AI techniques for the game of Go}.
\newblock Citeseer.

\bibitem[{Van~Gulick(2022)}]{sep-consciousness}
Robert Van~Gulick. 2022.
\newblock {Consciousness}.
\newblock In Edward~N. Zalta and Uri Nodelman, editors, \emph{The {Stanford}
  Encyclopedia of Philosophy}, {W}inter 2022 edition. Metaphysics Research Lab,
  Stanford University.

\bibitem[{Wang et~al.(2018)Wang, Singh, Michael, Hill, Levy, and
  Bowman}]{wang-etal-2018-glue}
Alex Wang, Amanpreet Singh, Julian Michael, Felix Hill, Omer Levy, and Samuel
  Bowman. 2018.
\newblock \href {https://doi.org/10.18653/v1/W18-5446} {{GLUE}: A multi-task
  benchmark and analysis platform for natural language understanding}.
\newblock In \emph{Proceedings of the 2018 {EMNLP} Workshop {B}lackbox{NLP}:
  Analyzing and Interpreting Neural Networks for {NLP}}, pages 353--355,
  Brussels, Belgium. Association for Computational Linguistics.

\bibitem[{Williams et~al.(2018)Williams, Nangia, and Bowman}]{N18-1101}
Adina Williams, Nikita Nangia, and Samuel Bowman. 2018.
\newblock \href {https://doi.org/10.18653/v1/N18-1101} {A broad-coverage
  challenge corpus for sentence understanding through inference}.
\newblock In \emph{Proceedings of the 2018 Conference of the North {A}merican
  Chapter of the Association for Computational Linguistics: Human Language
  Technologies, Volume 1 (Long Papers)}, pages 1112--1122, New Orleans,
  Louisiana. Association for Computational Linguistics.

\end{thebibliography}
\bibliographystyle{acl_natbib}

\end{document}